\documentclass[12pt,a4paper]{article}

\usepackage{float}
\usepackage{url}

\usepackage[title]{appendix}
\usepackage{hyperref}
\usepackage{amsmath}
\usepackage{amssymb}
\usepackage{mathtools}
\usepackage{graphicx}
\usepackage{threeparttable}
\usepackage{adjustbox,lipsum}
\usepackage{algorithm}
\usepackage[noend]{algpseudocode}

\usepackage[a4paper,left=2.5cm,right=2.5cm,top=2.5cm,bottom=2.5cm]{geometry}

\begin{document}
\title{Isolation forests: looking beyond tree depth}
\author{David Cortes}
\maketitle

\begin{abstract}
The isolation forest algorithm for outlier detection exploits a simple yet effective observation: if taking some multivariate data and making uniformly random cuts across the feature space recursively, it will take fewer such random cuts for an outlier to be left alone in a given subspace as compared to regular observations. The original idea proposed an outlier score based on the tree depth (number of random cuts) required for isolation, but experiments here show that using information about the size of the feature space taken and the number of points assigned to it can result in improved results in many situations without any modification to the tree structure, especially in the presence of categorical features.
\end{abstract} 

\section{Introduction}

The isolation forest algorithm (a.k.a. "iForest", \cite{iso}) proposed a very intuituitive idea for detecting outliers in multivariate data: choose some random dimension/variable/column uniformly at random and some value also uniformly at random within the range of that variable, then group all points according to whether they are greater or smaller than this threshold in the chosen dimension, and repeat the process across these two groups until each observation/point/row becomes isolated (is the only observation in a given group). Under this scheme, outliers or anomalies will on average become isolated in fewer splits/groupings/branchings/divisions than regular observations, and thus the number of splits/cuts that it takes to isolate an observation like this can be used as a scoring metric for outlier/anomaly detection. Since such a randomized procedure can present a lot of variation across different runs, the process is repeated many times using different sub-samples of the data and the results averaged to obtain a final score.

This simple idea has achieved good results across a variety of benchmarks (see e.g. \cite{genif} and \cite{gain}), and many subsequent works have proposed enhancements in the tree-building process for better outlier detection (e.g. \cite{ext}, \cite{sci}, \cite{gain}, among many others) while keeping the same isolation depth metric as criteria for producing outlier scores.

Isolation depth as a metric nevertheless does not capture the full effect of a random cut in the obtained data distributions: if data were uniformly distributed, a random cut choosing the middle point would be expected to leave half of the data points at each side, and observing a more uneven distribution than that would be more telling of outlierness for the smaller group, which isolation depth does not capture in full as it is only looking at the number of points without considering how much of the feature space it took to divide them.

This work introduces a density criterion as scoring metric based on the ratio between the fraction of the points that are sent to a given side of a split and the fraction of the feature space that is taken by that split, but in a slightly different way as in e.g. the density estimation trees from \cite{det}, which as will be shown, can lead to improved results in many situations compared to simple isolation depth without any changes in the tree building process or the model structure.

\section{Related work}

Using density in isolation trees or in decision trees in general is not a new concept. For example, \cite{det} proposed a method for estimating the density or likelihood of observations based on building symmetric decision trees which maximize a gain criterion at each step given by the density of points (viewed as points-to-volume) in the subspaces that define each node, and using a ratio of points-to-volume as a scoring metric which, while not originally intended for outlier detection, can nevertheless achieve reasonably good results as an outlier detector as shown in \cite{gain}.

Density as a splitting criterion for decision trees aimed at outlier detection has also been proposed in e.g. \cite{ocrf}, which used a gain criterion based on points-to-volume for building trees while employing isolation depth as its scoring metric for outlierness (it should be noted that higher-volume regions are defined by fewer splits than lower-volume regions, thus points falling there would have a shorter expected isolation depth).

Density as a scoring metric for outlierness in isolation trees was also proposed in \cite{genif}, while using a splitting criterion for building trees that indirectly tries to maximize density.

Metrics other than isolation depth or density have also been proposed for isolation trees - for example, \cite{rrcf} proposed a "Co-Displacement" metric for its random-cut trees that is based on the complexity that each point adds to the generated model, which is possible to calculate for their proposed decision trees as opposed to regular isolation trees because the variable to split at each node is chosen with a probability proportional to their ranges instead of uniformly at random.

A penalization for isolation depth was proposed in \cite{sci} based on determining a reasonable range for the splitting variable at each node, with observations falling outside that range at prediction time being penalized by preventing those splits from counting towards the isolation depth calculation.

\section{Depth and density}

The isolation forest or "iForest" algorithm presented in \cite{iso} proposed using isolation depth under random splits - that is, the number of uniformly random partitions that it takes to assign a point into a node/subspace in which there are no other points alongside - as a scoring metric for determining the degree of outlierness or inlierness of observations, based on the intuition that, if an observation has a value that does not match with the rest, a uniformly random split has a higher chance of sending it alone into a branch or assigning it to the branch with fewer points.

Since outliers can only be considered to be so if their isolation depth is shorter than expected for the amount of points that are being processed, the procedure might be stopped before achieving full isolation of all points if the number of splits is already exceeding the height of a balanced binary tree, and a remaining depth extrapolated for all points in that node according to the expected number of splits that it would take to isolate uniformly-random points with uniformly-random splits, leading to a very computationally-efficient procedure. As analyzed in \cite{gain}, the effect of stopping early is not too large compared to pursuing full isolation, while making the procedure much faster and the models much smaller.

Further works have provided a more theorical analysis of isolation depth, such as \cite{genif}, \cite{lsh}, or \cite{kern}, with \cite{genif} comparing it to density and \cite{lsh} to Manhattan (L1) distance with standardized variables.

While intuitive, the concept of isolation depth however might not capture the full information that a random cut in the feature space could provide: if a split threshold makes a 50\%-50\% split of the available range and this produces a 99\%-1\% split of the data points, then it would be reasonable to consider points in the smaller branch as less common than if a 99\%-1\% split of the range had produced the same result, assuming that these splits are uniformly random, but isolation depth would consider them as equally anomalous. If data were all uniformly-random and the process were repeated ad-infinitum, then there should be no mismatch between points-vs-range taken by splits, but data of interest typically has variables with different distributions and relationships.

In particular, if there are categorical features with many possible values and splits on categorical features are determined by assigning one category to a branch and the rest to the other, these splits will in expectation put few observations in one split, making them look as outliers from an isolation depth perspective even if the category that goes to one branch is relatively more frequent in the node than the other categories.

It would also be natural to use density - defined as number of points divided by volume unit - as a scoring metric if dealing with points assigned to delimited subspaces as an isolation tree does, but its calculation is not entirely straightforward.

For example, one could think of the feature space spanned by the data as a box defined by the maximums and minimums of each variable, and the terminal nodes being smaller boxes defined by the splits (upper and lower bounds) that lead to each node, bounded from outside by the larger box from the full data ranges, and from that calculate a volume metric for each box as the product of the lengths across each dimension that it spans, producing a score for each terminal node in the tree as "number of points that reach this node divided by the volume metric"; or one could alternatively think of the terminal nodes as being sub-boxes within this larger box, each taking a fraction of the available volume space and a fraction of the available points, and calculate a metric as "fraction of points reaching this node divided by fraction of the feature space used by it".

However, these calculations would not take into account that the distributions and available ranges of variables change as more splits are performed - that is, making a split on a given variable is likely to shorten the range of a different variable in the sub-nodes that it produces too, but the calculation above would still assume full ranges being available. One could also think of looking at the actual feature space used in a node by finding the convex hull that encloses the points (perhaps considering also the bounds from the splits) and calculating a volume metric from it, but this would result in a very slow procedure that might not be tractable for all inputs, and is not well defined for isolated points.

As yet another alternative, one could think of each side of a split as taking a fraction of the points and a fraction of the available range, and calculate a density metric as the ratio of the cumulated products of these two. This has the advantages of being easy to calculate while still accounting for the shifting distributions, and having an expected logarithmic value of 0 at each step for symmetric distributions (matching to a value of 1).

Following \cite{sci}, one could also think of adding this as a penalization or bonus to the isolation depth, with each successive split adding a depth plus a penalty, which in this case would be smoother than the range-based penalty from \cite{sci}.

This work proposes the following two additional metrics for determining outlier scores from isolation forest models (for details about the algorithm, see \cite{iso}) as alternatives to isolation depth and to standard "boxed" density:
\begin{itemize}
\item Adjusted depth:\\ \\
While passing down each node of the tree, instead of adding $+1$ to a cumulated depth with each successive split, add instead a standardized number between $(0, 2)$ and centered around $1$ which penalizes asymmetries between points taken vs. space taken:
$$
\text{Adj} = \frac{2}{1 + \frac{1}{2 r}}
$$
Where $r = \frac{
|\{ x_i \:\:|\:\: x_i \leq t  \}|
}{
|\mathcal{X}|
} /
\frac{t - \min{\{x_i\}}}{\max{\{x_i\}} - \min{\{x_i\}}}
$ for the left branch and
$r = \frac{
|\{ x_i \:\:|\:\: x_i > t  \}|
}{
|\mathcal{X}|
} /
\frac{\max{\{x_i\}} - t}{\max{\{x_i\}} - \min{\{x_i\}}}
$ for the right branch, with $\mathcal{X}$ being the set of points within the sub-sample of the training data that reaches a given node during the tree building procedure, and $x_i$ the values for each point across the splitting column from the parent node (i.e. $r$ is the ratio between points taken by a branch and space taken by a branch).

The expected value of this metric under uniformly random splits and uniformly random data is the same as for regular isolation depth, thus calculation of the standardized outlier score does not need to be changed any further from \cite{iso}.

For categorical variables with splits of the type $\{x_i = c\}$ and $\{x_i \neq c\}$, this metric can be calculated by taking the fraction of the available categories taken instead (with one branch always taking only one category).

This specific formula does not have any theoretical justification behind it other than producing symmetric deviations around 1 when one of the ratios is fixed at 50\%, and a different formula might do better.

Given its limited range and its distribution, this metric may be using multiplicativaly instead of additively.\\
\item Density:\\ \\
Instead of looking at isolation depth, look at the ratio betewen the fraction of points reaching a terminal node and the cumulated fraction of the available feature space taken by that node at each step.

This can be calculated by taking the $r$ criterion from the adjusted depth, but using it multiplicatively - i.e. start with a ratio of $1$ at the root node, then at each partition, calculate $r$ and multiply the current score by it. For a more numerically stable calculation in a software implementation, one can use the logarithmic identity $\prod_i^n z_i = \exp(\sum_i^n \log{z_i})$, and can calculate it using the fraction of the ranges separately from the fraction of the observations ($\frac{a}{b} = \exp(\log{a} - \log{b})$).

The expected value of the logarithm of $r$ at each step is $0$ for uniformly-random data and splits (corresponding to a value of $r$ of $1$), and thus the expected value of the logarithm of this metric at a terminal node under uniformly-random data and splits is also $0$.

A standardized outlier score could in theory be calculated using the same transformation as in \cite{iso} assuming an expected value of $1$ for the standardizing constant - however, unlike with the adjusted depth above, this score could suffer from numerical instabilities since the terminal ratio can grow infinitely large and thus an average across a large amount of trees will make all observations have a standardized score closer to zero than to one. Potential remedies are:
\begin{enumerate}
\item Cap the maximum value of this score at some reasonable threshold, such as $log_2{|\mathcal{X}|}$.
\item Take the median value of this metric across trees instead of the mean.
\item \emph{(recommended)} Take the geometric mean instead of the arithmetic mean across trees, as the logarithm of this terminal $r$ value would have a more symmetric distribution and this aggregation would thus be more appropriate.
\end{enumerate}

Given that the expected value for the logarithm of this metric under uniformly-random data and splits does not depend on the number of observations, it has the advantage of allowing to mix scores from trees that have varying numbers of observations (i.e. can use a sub-sample with different size for each tree), which is something that might produce better results as suggested in \cite{lsh} and could be helpful in a streaming data scenario.
\end{itemize}

\section{Experiments}

The scoring metrics proposed in the previous section (with mean tree density calculated by setting a ceiling on the values) were evaluated by building isolation forest models and variations thereof in public datasets for outlier detection, producing outlier scores with the obtained models and comparing them against the following alternatives:
\begin{itemize}
\item Isolation depth: simple isolation depth as proposed in \cite{iso}.
\item Penalized isolation depth: simple isolation depth, but adding a penalty at prediction time (not counting a given node as adding towards the depth) as proposed in \cite{sci} if the value of the splitting variable for an observation exceeds an acceptable range, calculated from the observations that reached a given node during tree building time as:
$$[\min{\{x_i\}} - (\max{\{x_i\}} - \min{\{x_i\}}),\:\:\: \max{\{x_i\}} + (\max{\{x_i\}} - \min{\{x_i\}})]$$
This penalization was not calculated for categorical variables.
\item Co-Displacement: a measure of how much including or excluding an observation increases or decreases the model complexity as proposed in \cite{rrcf} (only for splits in which the variables are chosen proportionally to their ranges).
\item Boxed Density: density calculated as the fraction of the points in the training sub-sample reaching a terminal node vs. the fraction of the "boxed" feature space within the training data that defines that node, without taking into consideration that ranges of non-splitting variables also shrink when delving deeper down a tree. Being a ratio, it would be more intuitive to aggregate it across trees by its geometric mean, which was also evaluated here. For the experiments, the maximum value of this metric at each node was capped at $2 \log_2{|\mathcal{X}|}$ when taking the arithmetic mean, and at $100$ when taking the geometric mean.
\item Adjusted Density: using the adjusted depth multiplicatively instead of additively.
\end{itemize}

The implementation of the models and metrics used here is made open source and freely available\footnote{\url{https://github.com/david-cortes/isotree}}.
\\ \\
The datasets used for comparisons are the following:
\begin{table}[H]
\centering
\caption {Datasets used for comparisons}
\begin{adjustbox}{max width=\textwidth}{\centering
\begin{tabular}{|r|c|c|c|c|}
 \hline
 \textbf{Dataset} & \textbf{Rows} & \textbf{Columns} & \textbf{Outliers} \\
 \hline
Arrhythmia & 452 & 274 & 15\% \\  \hline
Pima & 768 & 8 & 35\% \\  \hline
Musk & 3,062 & 166 & 3.17\% \\  \hline
Waveform-1 & 3,443 & 21 & 2.90\% \\  \hline
Thyroid & 3,772 & 6 & 2.47\% \\  \hline
SpamBase & 4,601 & 57 & 39.4\% \\  \hline
Wilt & 4,839 & 5 & 5.39\% \\  \hline
Satimage-2 & 5,803 & 36 & 1.22\% \\  \hline
Satellite & 6,435 & 36 & 32\% \\  \hline
Pendigits & 6,870 & 16 & 2.27\% \\  \hline
Annthyroid & 7,200 & 6 & 7.42\% \\  \hline
Mnist & 7,603 & 100 & 9.2\% \\  \hline
Shuttle & 49,097 & 9 & 7.15\% \\  \hline
ALOI & 50,000 & 27* & 3\% \\  \hline
KDDCup99 & 60,839 & 41** & 0.40\% \\  \hline
ForestCover & 286,048 & 10 & 0.9\% \\  \hline
HTTP & 567,498 & 3 & 0.39\% \\  \hline
\end{tabular}}\end{adjustbox}
\begin{tablenotes}
      \footnotesize \item * This dataset contains a categorical variable which was ignored.
      \footnotesize \item ** This dataset contains 3 categorical variables which were used by producing splits in which a single category goes to one branch.
\end{tablenotes}
\end{table}

These represent a variety of problem domains and a variety of outlier types for which isolation forests might or might not be good choices (see \cite{gain}). They were all taken from \cite{odds} and \cite{repo2}, using the original data without transformations and preferring the version from \cite{odds} when available, which differs from how many others (e.g. \cite{genif}) have performed their experiments.

Performance was evaluated by building isolation forest models and variations thereof with their recommended hyperparameters (which again differs from e.g. \cite{genif}) and calculating area under the receiver-operating characteristic curve (AUROC) and area under the precision-recall curve (AUPR), which might be a more appropriate metric for outlier detection (see \cite{aupr}). These metrics were calculated from the true outlier labels (which the models do not see) on the same data on which the models were fit, and the metrics were calculated 10 times for each experiment using different random seeds, taking the average of the AUROC and AUPR obtained across these 10 trials as the final metric to report.

\subsection{Isolation forest}

This is the original isolation forest model proposed in \cite{iso} (uniformly-random column choice, uniformly-random split threshold selection, 100 trees, sub-sampling 256 observations without replacement for each tree, maximum depth capped at 8). Note that all the metrics compared here are calculated from identical decision trees in each run, as the tree-building procedure is not impacted by the choice of scoring metric.

\begin{table}[H]
\centering
\caption {AUROC for isolation forest using different outlier scoring criteria}
\begin{adjustbox}{max width=\textwidth}{\centering
\begin{tabular}{|r|c|c|c|c|c|c|c|c|c|}
\hline
\textbf{Dataset} & \textbf{Depth} & \textbf{Penalized} & \textbf{B. Dens. (mean)} & \textbf{B. Dens. (geom. mean)} & \textbf{Adj. Dens.} & \textbf{Adj. Depth} & \textbf{Dens. (mean)} & \textbf{Dens. (median)} & \textbf{Dens. (geom. mean)} \\
 \hline
 
\textbf{Arrhythmia} & 0.7966 & \textbf{0.7968} & 0.6610 & 0.7185 & 0.7845 & 0.7945 & 0.7543 & 0.7515 & 0.7644 \\ \hline 
\textbf{Pima} & 0.6395 & 0.6393 & 0.5745 & 0.4777 & 0.6760 & 0.6509 & \textbf{0.6884} & 0.6857 & 0.6689 \\ \hline 
\textbf{Musk} & 0.9994 & 0.9993 & 0.4765 & 0.0557 & \textbf{1.0000} & 0.9997 & 0.9998 & 0.9997 & 0.9999 \\ \hline 
\textbf{Waveform-1} & 0.6684 & 0.6696 & 0.6074 & 0.5268 & 0.7310 & 0.6768 & \textbf{0.7692} & 0.7493 & 0.7615 \\ \hline 
\textbf{Thyroid} & 0.9796 & 0.9786 & 0.9618 & 0.9422 & 0.9814 & 0.9794 & \textbf{0.9853} & 0.9772 & 0.9809 \\ \hline 
\textbf{SpamBase} & 0.6403 & 0.6384 & 0.4790 & 0.3016 & 0.7126 & 0.6503 & 0.6185 & \textbf{0.7193} & 0.6403 \\ \hline 
\textbf{Wilt} & 0.4211 & 0.4240 & 0.3620 & 0.3441 & 0.4454 & 0.4259 & 0.4608 & \textbf{0.4622} & 0.4592 \\ \hline 
\textbf{Satimage-2} & 0.9932 & 0.9938 & 0.9606 & 0.9530 & 0.9962 & 0.9939 & 0.9951 & 0.9952 & \textbf{0.9963} \\ \hline 
\textbf{Satellite} & 0.7047 & 0.7056 & 0.6097 & 0.6083 & 0.8149 & 0.7211 & \textbf{0.8515} & 0.8451 & 0.8408 \\ \hline 
\textbf{Pendigits} & \textbf{0.9485} & 0.9467 & 0.8399 & 0.7351 & 0.9450 & 0.9474 & 0.9232 & 0.9090 & 0.9216 \\ \hline 
\textbf{Annthyroid} & 0.8480 & 0.8473 & 0.6919 & 0.6842 & 0.8949 & 0.8546 & 0.8972 & 0.8850 & \textbf{0.9100} \\ \hline 
\textbf{Mnist} & \textbf{0.7878} & 0.7865 & 0.5307 & 0.6634 & 0.7020 & 0.7733 & 0.6880 & 0.6767 & 0.7003 \\ \hline 
\textbf{Shuttle} & 0.9980 & \textbf{0.9981} & 0.9810 & 0.9067 & 0.9966 & 0.9981 & 0.9959 & 0.9951 & 0.9977 \\ \hline 
\textbf{ALOI} & 0.5424 & 0.5428 & 0.5409 & \textbf{0.5526} & 0.5327 & 0.5403 & 0.5288 & 0.5203 & 0.5299 \\ \hline 
\textbf{KDDCup99} & 0.9885 & 0.9888 & 0.9501 & 0.8840 & 0.9869 & \textbf{0.9890} & 0.9857 & 0.9848 & 0.9850 \\ \hline 
\textbf{ForestCover} & 0.8679 & 0.8721 & 0.8551 & \textbf{0.9493} & 0.7093 & 0.8559 & 0.5700 & 0.5793 & 0.6648 \\ \hline 
\textbf{HTTP} & 0.9999 & 0.9994 & 0.9971 & \textbf{1.0000} & 1.0000 & 0.9999 & 0.9999 & 0.9986 & 0.9999 \\ \hline \hline
\textbf{Geom. Mean} & 0.7908 & 0.7912 & 0.6801 & 0.5741 & \textbf{0.7981} & 0.7931 & 0.7841 & 0.7867 & 0.7923 \\ \hline
\end{tabular}}\end{adjustbox}
\end{table}

\begin{table}[H]
\centering
\caption {AUPR for isolation forest using different outlier scoring criteria}
\begin{adjustbox}{max width=\textwidth}{\centering
\begin{tabular}{|r|c|c|c|c|c|c|c|c|c|}
\hline
\textbf{Dataset} & \textbf{Depth} & \textbf{Penalized} & \textbf{B. Dens. (mean)} & \textbf{B. Dens. (geom. mean)} & \textbf{Adj. Dens.} & \textbf{Adj. Depth} & \textbf{Dens. (mean)} & \textbf{Dens. (median)} & \textbf{Dens. (geom. mean)} \\
 \hline
 
\textbf{Arrhythmia} & \textbf{0.4788} & 0.4778 & 0.3194 & 0.3012 & 0.4700 & 0.4749 & 0.4139 & 0.3996 & 0.4145 \\ \hline 
\textbf{Pima} & 0.4648 & 0.4659 & 0.4076 & 0.3468 & 0.5209 & 0.4766 & \textbf{0.5486} & 0.5378 & 0.5281 \\ \hline 
\textbf{Musk} & 0.9852 & 0.9837 & 0.0282 & 0.0172 & \textbf{0.9997} & 0.9932 & 0.9933 & 0.9927 & 0.9979 \\ \hline 
\textbf{Waveform-1} & 0.0571 & 0.0571 & 0.0423 & 0.0295 & 0.0647 & 0.0581 & 0.0714 & \textbf{0.0757} & 0.0703 \\ \hline 
\textbf{Thyroid} & 0.5727 & 0.5082 & 0.3531 & 0.4953 & 0.6314 & 0.5630 & \textbf{0.7041} & 0.4996 & 0.6250 \\ \hline 
\textbf{SpamBase} & 0.4737 & 0.4718 & 0.3583 & 0.2841 & 0.5475 & 0.4823 & 0.4623 & \textbf{0.5518} & 0.4757 \\ \hline 
\textbf{Wilt} & 0.0422 & 0.0424 & 0.0406 & 0.0373 & 0.0439 & 0.0425 & 0.0452 & \textbf{0.0454} & 0.0451 \\ \hline 
\textbf{Satimage-2} & 0.9197 & 0.9268 & 0.4424 & 0.3199 & 0.9466 & 0.9271 & 0.9440 & 0.9479 & \textbf{0.9540} \\ \hline 
\textbf{Satellite} & 0.6579 & 0.6521 & 0.6005 & 0.5534 & 0.7548 & 0.6712 & 0.7645 & \textbf{0.7806} & 0.7729 \\ \hline 
\textbf{Pendigits} & \textbf{0.3286} & 0.3077 & 0.1117 & 0.1025 & 0.2898 & 0.3114 & 0.2100 & 0.1915 & 0.2048 \\ \hline 
\textbf{Annthyroid} & 0.3039 & 0.2957 & 0.2330 & 0.2433 & 0.3549 & 0.3046 & \textbf{0.4239} & 0.3169 & 0.3872 \\ \hline 
\textbf{Mnist} & 0.2484 & 0.2462 & 0.1352 & \textbf{0.3049} & 0.1750 & 0.2351 & 0.1690 & 0.1619 & 0.1755 \\ \hline 
\textbf{Shuttle} & 0.9822 & 0.9720 & 0.8578 & 0.4558 & 0.9858 & 0.9841 & 0.9845 & 0.9849 & \textbf{0.9888} \\ \hline 
\textbf{ALOI} & 0.0330 & 0.0331 & \textbf{0.0345} & 0.0340 & 0.0319 & 0.0328 & 0.0319 & 0.0310 & 0.0319 \\ \hline 
\textbf{KDDCup99} & 0.3228 & \textbf{0.3375} & 0.1308 & 0.0558 & 0.3213 & 0.3132 & 0.2367 & 0.2093 & 0.2100 \\ \hline 
\textbf{ForestCover} & 0.0435 & 0.0440 & 0.0640 & \textbf{0.0947} & 0.0191 & 0.0389 & 0.0110 & 0.0127 & 0.0152 \\ \hline 
\textbf{HTTP} & 0.9854 & 0.8392 & 0.5649 & \textbf{0.9971} & 0.9932 & 0.9887 & 0.9861 & 0.7501 & 0.9914 \\ \hline \hline
\textbf{Geom. Mean} & \textbf{0.2949} & 0.2892 & 0.1724 & 0.1603 & 0.2865 & 0.2919 & 0.2694 & 0.2563 & 0.2696 \\ \hline
\end{tabular}}\end{adjustbox}
\end{table}

\begin{table}[H]
\centering
\caption {Percentage of wins in terms of AUROC of each metric over every other metric}
\begin{adjustbox}{max width=\textwidth}{\centering
\begin{tabular}{|r|c|c|c|c|c|c|c|c|c|}
\hline
\textbf{Wins over} & \textbf{Depth} & \textbf{Penalized} & \textbf{B. Dens. (mean)} & \textbf{B. Dens. (geom. mean)} & \textbf{Adj. Dens.} & \textbf{Adj. Depth} & \textbf{Dens. (mean)} & \textbf{Dens. (median)} & \textbf{Dens. (geom. mean)} \\
 \hline

\textbf{Depth} & - & 52.94\% & 0.00\% & 17.65\% & 58.82\% & 64.71\% & 47.06\% & 47.06\% & 52.94\% \\ \hline
\textbf{Penalized} & 47.06\% & - & 0.00\% & 17.65\% & 58.82\% & 70.59\% & 52.94\% & 47.06\% & 58.82\% \\ \hline
\textbf{B. Dens. (mean)} & 100.00\% & 100.00\% & - & 29.41\% & 88.24\% & 94.12\% & 88.24\% & 88.24\% & 88.24\% \\ \hline
\textbf{B. Dens. (geom. mean)} & 82.35\% & 82.35\% & 70.59\% & - & 82.35\% & 82.35\% & 82.35\% & 82.35\% & 82.35\% \\ \hline
\textbf{Adj. Dens.} & 41.18\% & 41.18\% & 11.76\% & 17.65\% & - & 41.18\% & 35.29\% & 29.41\% & 35.29\% \\ \hline
\textbf{Adj. Depth} & 35.29\% & 29.41\% & 5.88\% & 17.65\% & 58.82\% & - & 47.06\% & 41.18\% & 47.06\% \\ \hline
\textbf{Dens. (mean)} & 52.94\% & 47.06\% & 11.76\% & 17.65\% & 64.71\% & 52.94\% & - & 23.53\% & 58.82\% \\ \hline
\textbf{Dens. (median)} & 52.94\% & 52.94\% & 11.76\% & 17.65\% & 70.59\% & 58.82\% & 76.47\% & - & 76.47\% \\ \hline
\textbf{Dens. (geom. mean)} & 47.06\% & 41.18\% & 11.76\% & 17.65\% & 64.71\% & 52.94\% & 41.18\% & 23.53\% & - \\ \hline
\end{tabular}}\end{adjustbox}
\end{table}

\begin{table}[H]
\centering
\caption {Percentage of wins in terms of AUPR of each metric over every other metric}
\begin{adjustbox}{max width=\textwidth}{\centering
\begin{tabular}{|r|c|c|c|c|c|c|c|c|c|}
\hline
\textbf{Wins over} & \textbf{Depth} & \textbf{Penalized} & \textbf{B. Dens. (mean)} & \textbf{B. Dens. (geom. mean)} & \textbf{Adj. Dens.} & \textbf{Adj. Depth} & \textbf{Dens. (mean)} & \textbf{Dens. (median)} & \textbf{Dens. (geom. mean)} \\
 \hline

\textbf{Depth} & - & 41.18\% & 11.76\% & 23.53\% & 64.71\% & 58.82\% & 58.82\% & 52.94\% & 64.71\% \\ \hline
\textbf{Penalized} & 58.82\% & - & 11.76\% & 23.53\% & 64.71\% & 70.59\% & 58.82\% & 52.94\% & 64.71\% \\ \hline
\textbf{B. Dens. (mean)} & 88.24\% & 88.24\% & - & 29.41\% & 88.24\% & 88.24\% & 88.24\% & 88.24\% & 88.24\% \\ \hline
\textbf{B. Dens. (geom. mean)} & 76.47\% & 76.47\% & 70.59\% & - & 76.47\% & 76.47\% & 76.47\% & 76.47\% & 76.47\% \\ \hline
\textbf{Adj. Dens.} & 35.29\% & 35.29\% & 11.76\% & 23.53\% & - & 29.41\% & 35.29\% & 35.29\% & 47.06\% \\ \hline
\textbf{Adj. Depth} & 41.18\% & 29.41\% & 11.76\% & 23.53\% & 70.59\% & - & 52.94\% & 47.06\% & 58.82\% \\ \hline
\textbf{Dens. (mean)} & 41.18\% & 41.18\% & 11.76\% & 23.53\% & 64.71\% & 47.06\% & - & 41.18\% & 58.82\% \\ \hline
\textbf{Dens. (median)} & 47.06\% & 47.06\% & 11.76\% & 23.53\% & 64.71\% & 52.94\% & 58.82\% & - & 70.59\% \\ \hline
\textbf{Dens. (geom. mean)} & 35.29\% & 35.29\% & 11.76\% & 23.53\% & 52.94\% & 41.18\% & 41.18\% & 29.41\% & - \\ \hline
\end{tabular}}\end{adjustbox}
\end{table}

\subsection{Categorical variables}

Oftentimes, data of interest contains a mixture of numeric and categorical variables. In most of the public datasets used for outlier detection however, all or the vast majority of the variables are numeric.

In order to evaluate the impact of the scoring metric on models fit to data with categorical variables, the same datasets from the previous experiments were converted to categorical by discretizing each variable into 10 bins, with each bin taking a tenth of the range, and these discretized versions of the features treated as unordered categorical variables, creating splits by assigning one category to one branch and the rest to the other.

\begin{table}[H]
\centering
\caption {AUROC for isolation forest with discretized variables treated as categorical, using different outlier scoring criteria}
\begin{adjustbox}{max width=\textwidth}{\centering
\begin{tabular}{|r|c|c|c|c|c|c|c|c|}
\hline
\textbf{Dataset} & \textbf{Depth} & \textbf{B. Dens. (mean)} & \textbf{B. Dens. (geom. mean)} & \textbf{Adj. Dens.} & \textbf{Adj. Depth} & \textbf{Dens. (mean)} & \textbf{Dens. (median)} & \textbf{Dens. (geom. mean)} \\
 \hline
 
\textbf{Arrhythmia} & 0.7499 & 0.7632 & 0.7870 & 0.3039 & 0.7426 & 0.7826 & 0.7836 & \textbf{0.7871} \\ \hline 
\textbf{Pima} & 0.6741 & \textbf{0.6828} & 0.6663 & 0.4249 & 0.6652 & 0.6776 & 0.6764 & 0.6719 \\ \hline 
\textbf{Musk} & 0.8905 & 0.9926 & 0.9996 & 0.4598 & 0.8517 & \textbf{0.9998} & 0.9973 & 0.9990 \\ \hline 
\textbf{Waveform-1} & 0.5947 & 0.6673 & \textbf{0.6813} & 0.4293 & 0.5623 & 0.6586 & 0.6741 & 0.6666 \\ \hline 
\textbf{Thyroid} & 0.9604 & 0.9596 & 0.9622 & 0.1871 & 0.9564 & 0.9580 & \textbf{0.9641} & 0.9628 \\ \hline 
\textbf{SpamBase} & 0.6413 & 0.5844 & 0.6023 & 0.3876 & \textbf{0.6522} & 0.6012 & 0.6474 & 0.6244 \\ \hline 
\textbf{Wilt} & 0.3741 & 0.3494 & 0.3407 & \textbf{0.5605} & 0.3862 & 0.3589 & 0.3558 & 0.3582 \\ \hline 
\textbf{Satimage-2} & 0.9444 & 0.9872 & \textbf{0.9909} & 0.0724 & 0.7834 & 0.9759 & 0.9740 & 0.9836 \\ \hline 
\textbf{Satellite} & 0.7890 & 0.7404 & 0.7706 & 0.3871 & \textbf{0.7913} & 0.7857 & 0.7486 & 0.7734 \\ \hline 
\textbf{Pendigits} & \textbf{0.9272} & 0.9069 & 0.9218 & 0.3882 & 0.9216 & 0.9120 & 0.8953 & 0.9256 \\ \hline 
\textbf{Annthyroid} & 0.6347 & 0.6357 & \textbf{0.6393} & 0.4295 & 0.6386 & 0.6310 & 0.6274 & 0.6339 \\ \hline 
\textbf{Mnist} & 0.6843 & 0.6004 & 0.6241 & 0.4657 & \textbf{0.6997} & 0.6029 & 0.5972 & 0.6363 \\ \hline 
\textbf{Shuttle} & 0.9538 & \textbf{0.9753} & 0.9586 & 0.0600 & 0.9203 & 0.9675 & 0.9496 & 0.9572 \\ \hline 
\textbf{ALOI} & 0.5352 & 0.5377 & 0.5376 & 0.4751 & 0.5326 & 0.5362 & \textbf{0.5377} & 0.5355 \\ \hline 
\textbf{KDDCup99} & 0.9727 & 0.9741 & \textbf{0.9749} & 0.0643 & 0.9711 & 0.9736 & 0.9700 & 0.9730 \\ \hline 
\textbf{ForestCover} & 0.5360 & 0.6186 & \textbf{0.6768} & 0.3806 & 0.5013 & 0.6195 & 0.6137 & 0.6511 \\ \hline 
\textbf{HTTP} & \textbf{0.9995} & 0.9990 & 0.9990 & 0.3773 & 0.9953 & 0.9982 & 0.9943 & 0.9995 \\ \hline \hline
\textbf{Geom. Mean} & 0.7316 & 0.7356 & 0.7451 & 0.2866 & 0.7165 & 0.7401 & 0.7390 & \textbf{0.7468} \\ \hline
\end{tabular}}\end{adjustbox}
\end{table}

\begin{table}[H]
\centering
\caption {AUPR for isolation forest with discretized variables treated as categorical, using different outlier scoring criteria}
\begin{adjustbox}{max width=\textwidth}{\centering
\begin{tabular}{|r|c|c|c|c|c|c|c|c|}
\hline
\textbf{Dataset} & \textbf{Depth} & \textbf{B. Dens. (mean)} & \textbf{B. Dens. (geom. mean)} & \textbf{Adj. Dens.} & \textbf{Adj. Depth} & \textbf{Dens. (mean)} & \textbf{Dens. (median)} & \textbf{Dens. (geom. mean)} \\
 \hline
 
\textbf{Arrhythmia} & 0.3813 & 0.4211 & 0.4294 & 0.1004 & 0.3611 & 0.4220 & 0.4097 & \textbf{0.4303} \\ \hline 
\textbf{Pima} & \textbf{0.5500} & 0.5403 & 0.5334 & 0.3042 & 0.5440 & 0.5464 & 0.5462 & 0.5409 \\ \hline 
\textbf{Musk} & 0.3542 & 0.8462 & 0.9817 & 0.0273 & 0.3163 & \textbf{0.9941} & 0.9414 & 0.9740 \\ \hline 
\textbf{Waveform-1} & 0.0381 & 0.0585 & \textbf{0.0591} & 0.0252 & 0.0339 & 0.0519 & 0.0508 & 0.0538 \\ \hline 
\textbf{Thyroid} & 0.5936 & 0.5538 & 0.5802 & 0.0161 & 0.5429 & 0.5427 & 0.5308 & \textbf{0.5997} \\ \hline 
\textbf{SpamBase} & 0.4871 & 0.4519 & 0.4585 & 0.3255 & \textbf{0.5005} & 0.4531 & 0.4940 & 0.4743 \\ \hline 
\textbf{Wilt} & 0.0418 & 0.0403 & 0.0395 & \textbf{0.0604} & 0.0427 & 0.0408 & 0.0405 & 0.0406 \\ \hline 
\textbf{Satimage-2} & 0.1173 & \textbf{0.7074} & 0.6699 & 0.0065 & 0.0298 & 0.3297 & 0.2888 & 0.4128 \\ \hline 
\textbf{Satellite} & 0.6864 & 0.6340 & 0.6996 & 0.2627 & 0.6750 & \textbf{0.7330} & 0.7002 & 0.7099 \\ \hline 
\textbf{Pendigits} & \textbf{0.2958} & 0.2156 & 0.2526 & 0.0169 & 0.2633 & 0.1782 & 0.1934 & 0.2623 \\ \hline 
\textbf{Annthyroid} & 0.2186 & 0.1500 & 0.1733 & 0.0644 & \textbf{0.2636} & 0.2360 & 0.1587 & 0.2030 \\ \hline 
\textbf{Mnist} & 0.1683 & 0.1280 & 0.1482 & 0.0811 & \textbf{0.1722} & 0.1229 & 0.1171 & 0.1503 \\ \hline 
\textbf{Shuttle} & 0.5373 & \textbf{0.8180} & 0.6095 & 0.0436 & 0.3689 & 0.6961 & 0.5149 & 0.5784 \\ \hline 
\textbf{ALOI} & 0.0321 & \textbf{0.0332} & 0.0328 & 0.0291 & 0.0324 & 0.0327 & 0.0324 & 0.0324 \\ \hline 
\textbf{KDDCup99} & 0.1994 & \textbf{0.2637} & 0.1837 & 0.0025 & 0.2108 & 0.2162 & 0.1872 & 0.1915 \\ \hline 
\textbf{ForestCover} & 0.0113 & 0.0177 & \textbf{0.0185} & 0.0077 & 0.0098 & 0.0173 & 0.0150 & 0.0184 \\ \hline 
\textbf{HTTP} & 0.9155 & 0.8047 & 0.7941 & 0.0065 & 0.4631 & \textbf{0.9549} & 0.4591 & 0.9041 \\ \hline \hline
\textbf{Geom. Mean} & 0.1976 & 0.2358 & \textbf{0.2366} & 0.0348 & 0.1679 & 0.2280 & 0.2036 & 0.2346 \\ \hline 
\end{tabular}}\end{adjustbox}
\end{table}

\begin{table}[H]
\centering
\caption {Percentage of wins in terms of AUROC of each metric over every other metric}
\begin{adjustbox}{max width=\textwidth}{\centering
\begin{tabular}{|r|c|c|c|c|c|c|c|c|}
\hline
\textbf{Wins over} & \textbf{Depth} & \textbf{B. Dens. (mean)} & \textbf{B. Dens. (geom. mean)} & \textbf{Adj. Dens.} & \textbf{Adj. Depth} & \textbf{Dens. (mean)} & \textbf{Dens. (median)} & \textbf{Dens. (geom. mean)} \\
 \hline

\textbf{Depth} & - & 58.82\% & 58.82\% & 5.88\% & 29.41\% & 52.94\% & 52.94\% & 52.94\% \\ \hline
\textbf{B. Dens. (mean)} & 41.18\% & - & 70.59\% & 5.88\% & 35.29\% & 47.06\% & 47.06\% & 58.82\% \\ \hline
\textbf{B. Dens. (geom. mean)} & 41.18\% & 29.41\% & - & 5.88\% & 23.53\% & 29.41\% & 29.41\% & 52.94\% \\ \hline
\textbf{Adj. Dens.} & 94.12\% & 94.12\% & 94.12\% & - & 94.12\% & 94.12\% & 94.12\% & 94.12\% \\ \hline
\textbf{Adj. Depth} & 70.59\% & 64.71\% & 76.47\% & 5.88\% & - & 64.71\% & 52.94\% & 70.59\% \\ \hline
\textbf{Dens. (mean)} & 47.06\% & 52.94\% & 70.59\% & 5.88\% & 35.29\% & - & 29.41\% & 58.82\% \\ \hline
\textbf{Dens. (median)} & 47.06\% & 52.94\% & 70.59\% & 5.88\% & 47.06\% & 70.59\% & - & 70.59\% \\ \hline
\textbf{Dens. (geom. mean)} & 47.06\% & 41.18\% & 47.06\% & 5.88\% & 29.41\% & 41.18\% & 29.41\% & - \\ \hline
\end{tabular}}\end{adjustbox}
\end{table}

\begin{table}[H]
\centering
\caption {Percentage of wins in terms of AUPR of each metric over every other metric}
\begin{adjustbox}{max width=\textwidth}{\centering
\begin{tabular}{|r|c|c|c|c|c|c|c|c|}
\hline
\textbf{Wins over} & \textbf{Depth} & \textbf{B. Dens. (mean)} & \textbf{B. Dens. (geom. mean)} & \textbf{Adj. Dens.} & \textbf{Adj. Depth} & \textbf{Dens. (mean)} & \textbf{Dens. (median)} & \textbf{Dens. (geom. mean)} \\
 \hline

\textbf{Depth} & - & 47.06\% & 47.06\% & 5.88\% & 35.29\% & 64.71\% & 47.06\% & 52.94\% \\ \hline
\textbf{B. Dens. (mean)} & 52.94\% & - & 58.82\% & 5.88\% & 41.18\% & 47.06\% & 35.29\% & 70.59\% \\ \hline
\textbf{B. Dens. (geom. mean)} & 52.94\% & 41.18\% & - & 5.88\% & 41.18\% & 47.06\% & 29.41\% & 64.71\% \\ \hline
\textbf{Adj. Dens.} & 94.12\% & 94.12\% & 94.12\% & - & 94.12\% & 94.12\% & 94.12\% & 94.12\% \\ \hline
\textbf{Adj. Depth} & 64.71\% & 58.82\% & 58.82\% & 5.88\% & - & 64.71\% & 52.94\% & 58.82\% \\ \hline
\textbf{Dens. (mean)} & 35.29\% & 52.94\% & 52.94\% & 5.88\% & 35.29\% & - & 11.76\% & 47.06\% \\ \hline
\textbf{Dens. (median)} & 52.94\% & 64.71\% & 70.59\% & 5.88\% & 47.06\% & 88.24\% & - & 88.24\% \\ \hline
\textbf{Dens. (geom. mean)} & 47.06\% & 29.41\% & 35.29\% & 5.88\% & 41.18\% & 52.94\% & 11.76\% & - \\ \hline
\end{tabular}}\end{adjustbox}
\end{table}

\subsection{Other variations}

These same metrics can also be calculated for other variations of isolation forests that build the trees differently, as they are independent of the tree building procedure.

\subsubsection{Robust random cut forest}

This is the model from \cite{rrcf} which chooses the variable to split at each node with a probability proportional to the range spanned by each variable among the observations that reached the splitting node during training time.

This procedure was intended in \cite{rrcf} for a streaming data scenario. Since it was used here for static data, it was done a bit differently, by sub-sampling 256 observations at random for each tree and building 100 trees just like in \cite{iso}, but without a depth limit in order to better match \cite{rrcf}.

Note that the "Co-Displacement" metric presented here was calculated with models built using a different software implementation\footnote{\url{https://github.com/kLabUM/rrcf}}, thus the trees are not the exact same ones as for the other metrics, but this should not make much of a difference when repeating the experiment across many randomized trials.

\begin{table}[H]
\centering
\caption {AUROC for robust random-cut forest using different outlier scoring criteria}
\begin{adjustbox}{max width=\textwidth}{\centering
\begin{tabular}{|r|c|c|c|c|c|c|c|c|c|c|}
\hline
\textbf{Dataset} & \textbf{Depth} & \textbf{Penalized} & \textbf{B. Dens. (mean)} & \textbf{B. Dens. (geom. mean)} & \textbf{Adj. Dens.} & \textbf{Adj. Depth} & \textbf{Dens. (mean)} & \textbf{Dens. (median)} & \textbf{Dens. (geom. mean)} & \textbf{CoDisp.} \\
 \hline

\textbf{Arrhythmia} & 0.8089 & 0.8082 & 0.5703 & 0.2570 & 0.8010 & 0.8099 & 0.8104 & 0.8113 & \textbf{0.8136} & 0.7844 \\ \hline 
\textbf{Pima} & 0.6347 & 0.6332 & 0.4179 & 0.4021 & \textbf{0.6389} & 0.6355 & 0.5942 & 0.5926 & 0.5909 & 0.5949 \\ \hline 
\textbf{Musk} & 1.0000 & 1.0000 & 0.0000 & 0.0000 & 0.9999 & 1.0000 & 1.0000 & 1.0000 & 1.0000 & 1.0000 \\ \hline 
\textbf{Waveform-1} & 0.7268 & 0.7312 & 0.1158 & 0.1039 & 0.7330 & 0.7345 & 0.8095 & \textbf{0.8214} & 0.8150 & 0.6983 \\ \hline 
\textbf{Thyroid} & 0.9550 & 0.9540 & 0.0756 & 0.5262 & 0.9491 & 0.9548 & 0.9554 & 0.9453 & 0.9563 & \textbf{0.9570} \\ \hline 
\textbf{SpamBase} & 0.7226 & 0.7223 & 0.7072 & 0.2283 & \textbf{0.7559} & 0.7292 & 0.7143 & 0.7553 & 0.7251 & 0.5772 \\ \hline 
\textbf{Wilt} & 0.4954 & 0.5017 & 0.3779 & 0.2636 & 0.4940 & 0.5000 & 0.5487 & \textbf{0.5497} & 0.5494 & 0.5451 \\ \hline 
\textbf{Satimage-2} & 0.9954 & 0.9961 & 0.0241 & 0.9102 & 0.9960 & 0.9960 & 0.9980 & \textbf{0.9984} & 0.9983 & 0.9961 \\ \hline 
\textbf{Satellite} & 0.7472 & 0.7472 & 0.3816 & 0.5827 & 0.7629 & 0.7564 & 0.8300 & 0.8210 & \textbf{0.8308} & 0.7112 \\ \hline 
\textbf{Pendigits} & 0.9532 & 0.9502 & 0.0897 & 0.2066 & 0.9485 & \textbf{0.9550} & 0.9444 & 0.9379 & 0.9474 & 0.9267 \\ \hline 
\textbf{Annthyroid} & 0.7163 & 0.7177 & 0.2131 & 0.3848 & 0.7161 & 0.7204 & 0.7532 & 0.7408 & \textbf{0.7739} & 0.7503 \\ \hline 
\textbf{Mnist} & \textbf{0.8212} & 0.8202 & 0.4962 & 0.5172 & 0.7943 & 0.8088 & 0.7066 & 0.7130 & 0.7211 & 0.7962 \\ \hline 
\textbf{Shuttle} & 0.9935 & 0.9934 & 0.0942 & 0.0536 & 0.9907 & 0.9938 & 0.9952 & 0.9927 & \textbf{0.9953} & 0.9842 \\ \hline 
\textbf{ALOI} & 0.5470 & 0.5491 & 0.4489 & 0.4753 & 0.5374 & 0.5445 & 0.5369 & 0.5300 & 0.5354 & \textbf{0.5635} \\ \hline 
\textbf{ForestCover} & 0.7703 & 0.7822 & 0.8315 & \textbf{0.9800} & 0.6176 & 0.7503 & 0.5086 & 0.5081 & 0.5688 & 0.9610 \\ \hline 
\textbf{HTTP} & 0.9957 & 0.9963 & 0.9955 & \textbf{0.9999} & 0.9910 & 0.9960 & 0.9962 & 0.9954 & 0.9950 & 0.9972 \\ \hline \hline
\textbf{Geom. Mean} & 0.7882 & \textbf{0.7897} & 0.1219 & 0.1964 & 0.7773 & 0.7884 & 0.7736 & 0.7744 & 0.7825 & 0.7842 \\ \hline
\end{tabular}}\end{adjustbox}
\end{table}

\subsubsection{Extended isolation forest}

This is the model proposed in \cite{ext} which makes splits according to random hyperplanes (using more than 1 variable) instead of using axis-parallel splits.

The boxed density calculation here would be more problematic, as each non-terminal node uses a different hyperplane - one could think of creating a transformed set of features consisting of all the linear combinations used for the splits defining an enclosing box and sub-boxes, but this might not have the same theoretical justification. For simplicity purposes, the "boxed density" calculation was ommited. As well, the KDDCup99 dataset was ommited due to having a categorical variable.

\begin{table}[H]
\centering
\caption {AUROC for extended isolation forest using different outlier scoring criteria}
\begin{adjustbox}{max width=\textwidth}{\centering
\begin{tabular}{|r|c|c|c|c|c|c|c|}
\hline
\textbf{Dataset} & \textbf{Depth} & \textbf{Penalized} & \textbf{Adj. Dens.} & \textbf{Adj. Depth} & \textbf{Dens. (mean)} & \textbf{Dens. (median)} & \textbf{Dens. (geom. mean)} \\
 \hline
 
\textbf{Arrhythmia} & 0.8153 & 0.8152 & 0.8130 & 0.8143 & 0.8114 & \textbf{0.8157} & 0.8031 \\ \hline 
\textbf{Pima} & \textbf{0.6584} & 0.6574 & 0.6505 & 0.6581 & 0.6582 & 0.6561 & 0.6359 \\ \hline 
\textbf{Musk} & 1.0000 & 1.0000 & 1.0000 & 1.0000 & 1.0000 & 1.0000 & 1.0000 \\ \hline 
\textbf{Waveform-1} & 0.7618 & 0.7630 & 0.8250 & 0.7721 & 0.8485 & \textbf{0.8519} & 0.8488 \\ \hline 
\textbf{Thyroid} & \textbf{0.9562} & 0.9559 & 0.9498 & 0.9561 & 0.9357 & 0.9438 & 0.9492 \\ \hline 
\textbf{SpamBase} & 0.6112 & 0.6102 & 0.6540 & 0.6178 & 0.6407 & \textbf{0.6963} & 0.6561 \\ \hline 
\textbf{Wilt} & 0.5183 & 0.5220 & 0.5594 & 0.5268 & 0.5739 & 0.5750 & \textbf{0.5795} \\ \hline 
\textbf{Satimage-2} & 0.9946 & 0.9951 & 0.9971 & 0.9953 & 0.9964 & \textbf{0.9978} & 0.9978 \\ \hline 
\textbf{Satellite} & 0.6941 & 0.6943 & 0.7724 & 0.7042 & \textbf{0.8068} & 0.7926 & 0.7914 \\ \hline 
\textbf{Pendigits} & 0.9505 & 0.9486 & 0.9521 & \textbf{0.9524} & 0.9391 & 0.9377 & 0.9454 \\ \hline 
\textbf{Annthyroid} & 0.7226 & 0.7246 & 0.7613 & 0.7300 & 0.7688 & 0.7592 & \textbf{0.7761} \\ \hline 
\textbf{Mnist} & \textbf{0.8236} & 0.8230 & 0.7678 & 0.8172 & 0.7660 & 0.7611 & 0.7544 \\ \hline 
\textbf{Shuttle} & 0.9976 & 0.9971 & 0.9970 & \textbf{0.9977} & 0.9964 & 0.9961 & 0.9974 \\ \hline 
\textbf{ALOI} & 0.5408 & \textbf{0.5414} & 0.5293 & 0.5387 & 0.5302 & 0.5283 & 0.5320 \\ \hline 
\textbf{ForestCover} & 0.8949 & \textbf{0.9004} & 0.8049 & 0.8933 & 0.6359 & 0.6437 & 0.7645 \\ \hline 
\textbf{HTTP} & 0.9987 & 0.9990 & \textbf{0.9990} & 0.9985 & 0.9927 & 0.9955 & 0.9953 \\ \hline \hline
\textbf{Geom. Mean} & 0.7906 & 0.7913 & 0.7990 & 0.7932 & 0.7902 & 0.7939 & \textbf{0.7990} \\ \hline 
\end{tabular}}\end{adjustbox}
\end{table}

\subsubsection{Split-criterion iForest}

This model as proposed in \cite{sci} makes non-random splits that maximize an averaged information gain criterion on randomly-generated hyperplanes, evaluating several random hyperplanes to choose the best one at each split. Just like for \textsc{EIF} above, boxed density and KDDCup99 were ommited.

\begin{table}[H]
\centering
\caption {AUROC for split-criterion isolation forest using different outlier scoring criteria}
\begin{adjustbox}{max width=\textwidth}{\centering
\begin{tabular}{|r|c|c|c|c|c|c|c|}
\hline
\textbf{Dataset} & \textbf{Depth} & \textbf{Penalized} & \textbf{Adj. Dens.} & \textbf{Adj. Depth} & \textbf{Dens. (mean)} & \textbf{Dens. (median)} & \textbf{Dens. (geom. mean)} \\
 \hline
 
\textbf{Arrhythmia} & 0.7016 & 0.7015 & 0.7141 & 0.7051 & 0.6953 & 0.6949 & \textbf{0.7182} \\ \hline 
\textbf{Pima} & 0.6141 & 0.6138 & 0.5702 & \textbf{0.6194} & 0.5562 & 0.5555 & 0.5622 \\ \hline 
\textbf{Musk} & 1.0000 & 1.0000 & 1.0000 & 1.0000 & 1.0000 & 1.0000 & 1.0000 \\ \hline 
\textbf{Waveform-1} & 0.6959 & 0.6919 & 0.6960 & 0.6964 & \textbf{0.7051} & 0.6804 & 0.7031 \\ \hline 
\textbf{Thyroid} & 0.9724 & \textbf{0.9736} & 0.9723 & 0.9724 & 0.9713 & 0.9685 & 0.9709 \\ \hline 
\textbf{SpamBase} & 0.4550 & 0.4502 & 0.4563 & 0.4550 & 0.4552 & \textbf{0.4886} & 0.4566 \\ \hline 
\textbf{Wilt} & 0.4900 & 0.4875 & 0.4901 & 0.4901 & 0.4899 & 0.4608 & \textbf{0.4903} \\ \hline 
\textbf{Satimage-2} & 0.9946 & 0.9946 & 0.9977 & 0.9952 & 0.9949 & 0.9917 & \textbf{0.9982} \\ \hline 
\textbf{Satellite} & 0.6158 & 0.6182 & 0.7061 & 0.6236 & 0.6994 & 0.7017 & \textbf{0.7111} \\ \hline 
\textbf{Pendigits} & 0.9795 & 0.9811 & \textbf{0.9818} & 0.9814 & 0.9759 & 0.9732 & 0.9753 \\ \hline 
\textbf{Annthyroid} & 0.8010 & 0.8004 & 0.8002 & 0.8009 & 0.7953 & \textbf{0.8033} & 0.8000 \\ \hline 
\textbf{Mnist} & \textbf{0.8486} & 0.8481 & 0.7705 & 0.8416 & 0.8023 & 0.6622 & 0.7686 \\ \hline 
\textbf{Shuttle} & 0.9982 & 0.9979 & 0.9989 & 0.9981 & \textbf{0.9993} & 0.9975 & 0.9975 \\ \hline 
\textbf{ALOI} & 0.5199 & \textbf{0.5206} & 0.5196 & 0.5199 & 0.5199 & 0.5168 & 0.5199 \\ \hline 
\textbf{ForestCover} & 0.6756 & 0.6774 & 0.7629 & 0.6905 & 0.6879 & 0.6202 & \textbf{0.7872} \\ \hline 
\textbf{HTTP} & 0.9999 & 0.9995 & \textbf{0.9999} & 0.9999 & 0.9998 & 0.9947 & 0.9952 \\ \hline \hline
\textbf{Geom. Mean} & 0.7465 & 0.7458 & 0.7517 & 0.7485 & 0.7457 & 0.7304 & \textbf{0.7529} \\ \hline
\end{tabular}}\end{adjustbox}
\end{table}

\subsubsection{Fair-cut forest}

This model as proposed in \cite{gain} makes non-random splits that maximize a pooled information gain criterion (which produces very different splits from maximizing an averaged information gain criterion) on randomly-generated hyperplanes, but evaluating only one such hyperplane at each node. Just like for \textsc{SCiF} and \textsc{EIF}, boxed density and KDDCup99 were ommited.

\begin{table}[H]
\centering
\caption {AUROC for fair-cut forest using different outlier scoring criteria}
\begin{adjustbox}{max width=\textwidth}{\centering
\begin{tabular}{|r|c|c|c|c|c|c|c|}
\hline
\textbf{Dataset} & \textbf{Depth} & \textbf{Penalized} & \textbf{Adj. Dens.} & \textbf{Adj. Depth} & \textbf{Dens. (mean)} & \textbf{Dens. (median)} & \textbf{Dens. (geom. mean)} \\
 \hline
 
\textbf{Arrhythmia} & 0.8087 & \textbf{0.8095} & 0.7985 & 0.8076 & 0.7999 & 0.8028 & 0.7989 \\ \hline 
\textbf{Pima} & \textbf{0.7337} & 0.7302 & 0.7333 & 0.7313 & 0.7252 & 0.7240 & 0.7183 \\ \hline 
\textbf{Musk} & 1.0000 & 1.0000 & 1.0000 & 1.0000 & 1.0000 & 1.0000 & 1.0000 \\ \hline 
\textbf{Waveform-1} & 0.8489 & 0.8393 & 0.8528 & 0.8576 & 0.8566 & 0.8753 & \textbf{0.8769} \\ \hline 
\textbf{Thyroid} & 0.9769 & \textbf{0.9769} & 0.9747 & 0.9752 & 0.9697 & 0.9639 & 0.9681 \\ \hline 
\textbf{SpamBase} & 0.7110 & 0.7098 & \textbf{0.7428} & 0.7130 & 0.6428 & 0.7135 & 0.6475 \\ \hline 
\textbf{Wilt} & 0.6738 & \textbf{0.6890} & 0.6228 & 0.6551 & 0.5773 & 0.5823 & 0.5739 \\ \hline 
\textbf{Satimage-2} & 0.9989 & \textbf{0.9990} & 0.9980 & 0.9987 & 0.9964 & 0.9978 & 0.9977 \\ \hline 
\textbf{Satellite} & \textbf{0.8367} & 0.8211 & 0.8345 & 0.8357 & 0.8264 & 0.8268 & 0.8248 \\ \hline 
\textbf{Pendigits} & \textbf{0.9626} & 0.9569 & 0.9517 & 0.9611 & 0.9499 & 0.9444 & 0.9515 \\ \hline 
\textbf{Annthyroid} & 0.8643 & \textbf{0.8752} & 0.8631 & 0.8610 & 0.8479 & 0.8359 & 0.8499 \\ \hline 
\textbf{Mnist} & \textbf{0.8112} & 0.8079 & 0.7964 & 0.8077 & 0.7821 & 0.7848 & 0.7935 \\ \hline 
\textbf{Shuttle} & \textbf{0.9979} & 0.9971 & 0.9960 & 0.9978 & 0.9958 & 0.9960 & 0.9971 \\ \hline 
\textbf{ALOI} & 0.5282 & 0.5303 & 0.5088 & 0.5281 & \textbf{0.5397} & 0.5300 & 0.5390 \\ \hline 
\textbf{ForestCover} & 0.5406 & \textbf{0.6358} & 0.5129 & 0.5364 & 0.5067 & 0.5216 & 0.5244 \\ \hline 
\textbf{HTTP} & 1.0000 & 0.9994 & 1.0000 & 1.0000 & 1.0000 & 0.9961 & 1.0000 \\ \hline \hline
\textbf{Geom. Mean} & 0.8150 & \textbf{0.8229} & 0.8062 & 0.8129 & 0.7946 & 0.8007 & 0.7978 \\ \hline
\end{tabular}}\end{adjustbox}
\end{table}

\section{Discussion}

From these experiments, no scoring metric is dominant over the others across all datasets, and a relative ranking of one metric over another would look different if evaluated according to ROC or PR curves. Some datasets present large performance differences between metrics (e.g. "SpamBase", "ForestCover") while others present little difference (e.g. "Arrythmia", "Pendigits").

The "boxed density", while perhaps very intuitive as a potential metric, in some cases fares so badly that using its inverse would produce a better ranking of anomalies which would be competitive against the other metrics, but in the absence of labels, it's not clear when to use it in ascending or descending order, and in a theorical sense, it would not be logical to consider denser subspaces as more anomalous than sparse ones. It also suffers from numerical instability as in some splits this metric can reach very large values which skew the final calculation, and a cap needs to be imposed. The exact value to use for capping is unclear however.

From the depth-based metrics, the "adjusted depth" proposed here more often than not provides better results than the regular or the penalized depths in the original isolation forest model, although the differences are not large.

From the density-based metrics, the density proposed here aggregated as a geometric mean across trees looks to be the best-performing one, and seems to be especially well suited to splits on categorical features. Unlike the "boxed density" metric, it does not seem to suffer from numerical instability issues in the datasets experimented with, although setting a maximum value such as $\log_2{|\mathcal{X}|}$ would make its distribution ranges very different in some datasets. These density-based metrics do not seem to perform as well under non-uniformly-random splits, which is reasonable to expect considering that e.g. the gain criterion from \textsc{FCF} already takes into consideration the number of observations and distributions under each branch. In some datasets, and in particular in the "ForestCover" dataset, many density-based metrics result in a very significant performance loss compared to depth-based metrics, even though the best performance was achieved with the "boxed density" metric. As found in \cite{gain}, tree-based methods might not be good choices for such types of outliers, regardless of the scoring metric used.

\section{A look at distributions}

Compared to isolation depth, the metrics introduced here present a larger degree of variation across trees:

\begin{figure}[H]
\centerline{\includegraphics[totalheight=5cm]{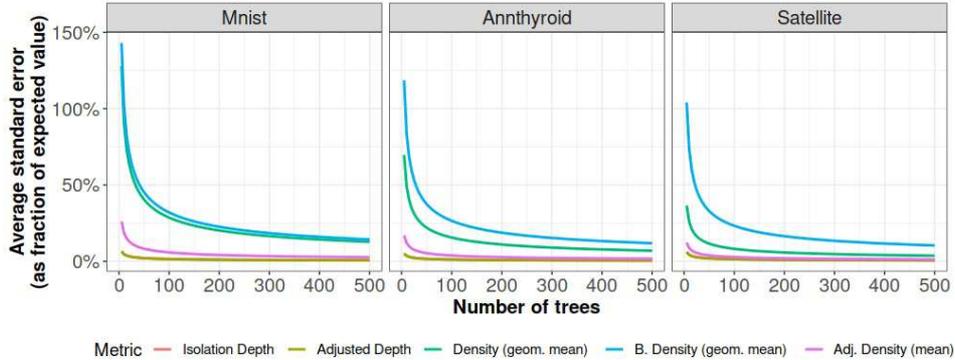}}
    \caption{Average standard error across observations of metric aggregation ("Isolation Depth" and "Adjusted Depth" share roughly the same line)}
    \label{fig:verticalcell1}
\end{figure}

In the "Mnist" dataset, calculating average isolation depth for an observation using 100 trees leads to an average standard error (as percentage of the expected value) among observations of around 1.5\%, but by 500 trees, this level of relative standard error is not yet achieved for any of the density-based metrics - the least variable, "adjusted density", still presenting an average standard error of around around 2.6\% at 500 trees.

However, when viewed in terms of achieved AUROC, this larger variance does not lead to performance degradataion for most metrics. In some datasets such as "Annthyroid", converge is similar for all metrics, while in others such as "Satellite", it does take more trees for density-based metrics to converge, but by 500 trees, convergence is already achieved despite what the standard error plot would suggest:

\begin{figure}[H]
\centerline{\includegraphics[totalheight=5cm]{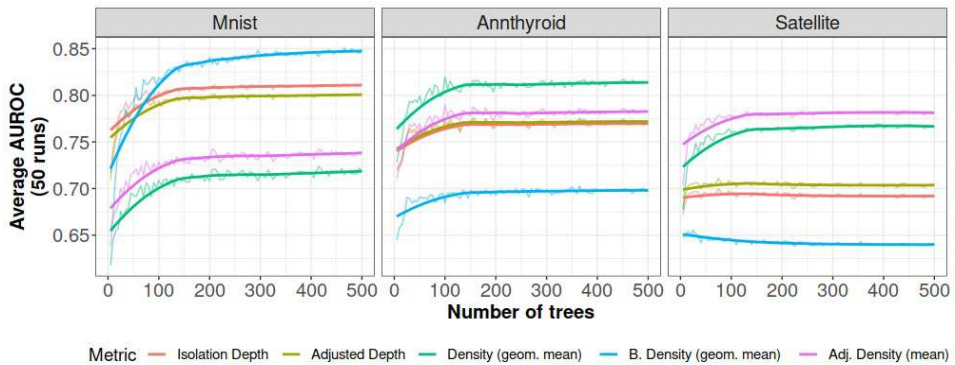}}
    \caption{AUROC by number of trees}
    \label{fig:verticalcell1}
\end{figure}

What these plots nonetheless suggest, is that 100 trees as recommended in \cite{iso} is not enough for any of the metrics, since even when using isolation depth, better AUROC in many datasets can be achieved by simply adding more trees. A safer choice would be 200 trees, increasing this recommendation to 300 trees for the "boxed density" metric.

From a different perspective, when observing the distributions of these density-based metrics across observations instead of across trees, they present more variability than the other metrics, and their distribution ranges have larger variations between datasets:

\begin{figure}[H]
\centerline{\includegraphics[totalheight=10cm]{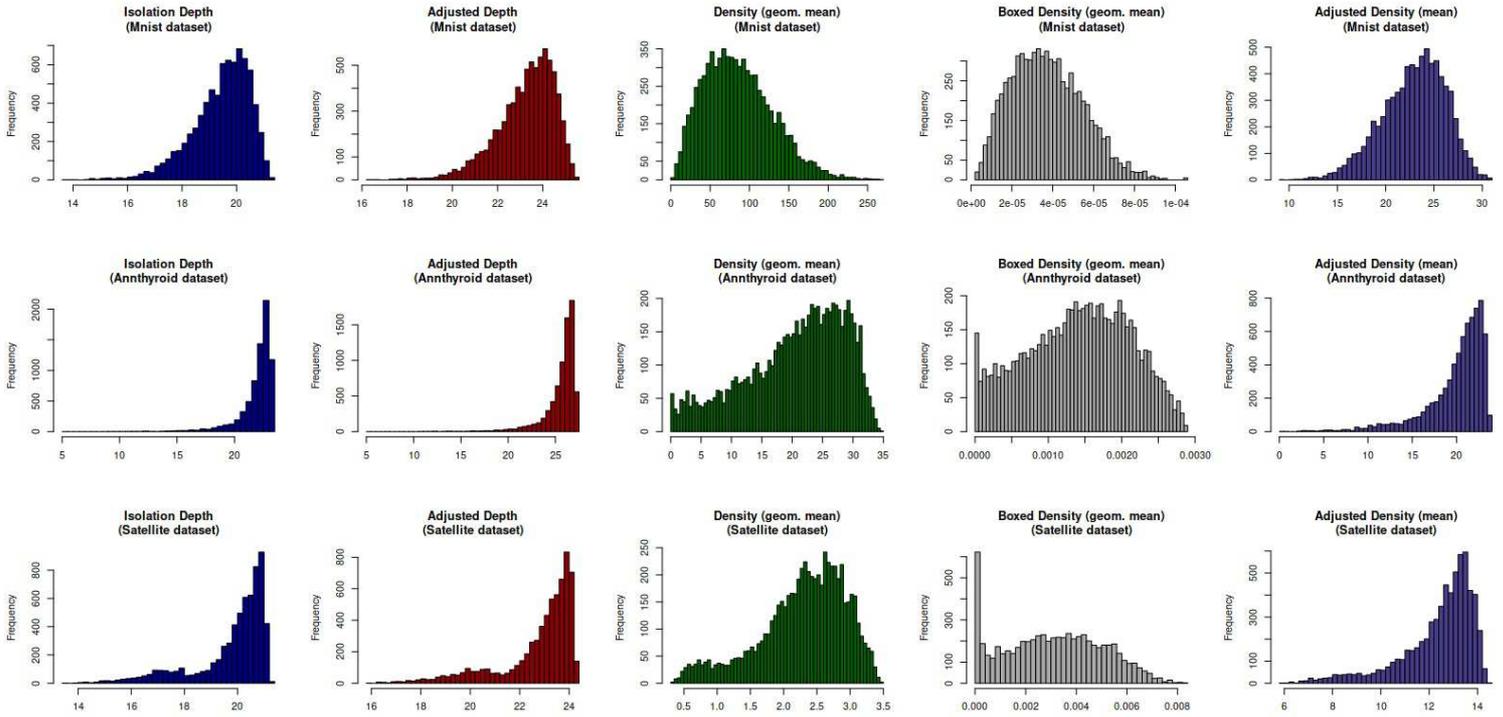}}
    \caption{Distribution of resulting aggregated metric (across trees) among observations}
    \label{fig:verticalcell1}
\end{figure}

If a standardized outlier score were to be calculated through the same transformation as for isolation depth, then the distribution of these scores would look different from those of depth-based metrics, with most of the observations looking as strong inliers or strong outliers, and with a much larger difference between a few outliers and the rest:
\begin{figure}[H]
\centerline{\includegraphics[totalheight=9cm]{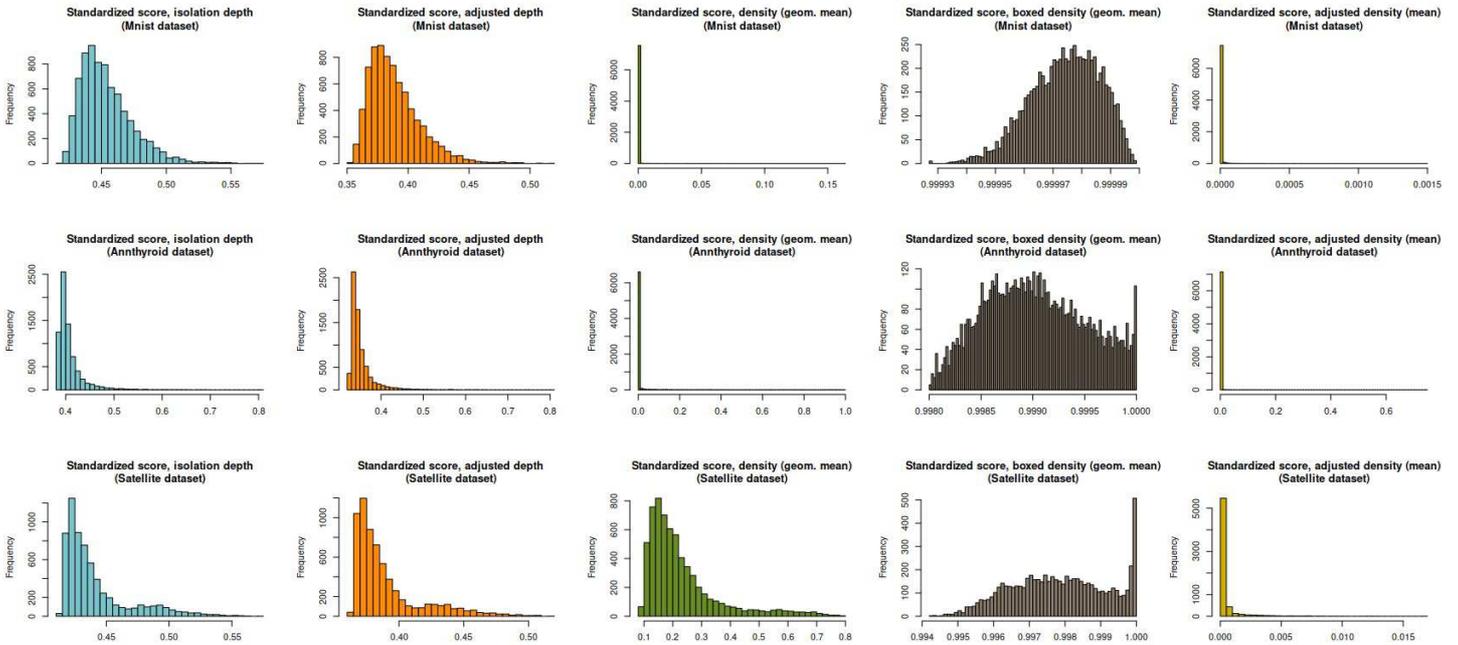}}
    \caption{Distribution of standardized outlier score among observations if the same transformation from \cite{iso} were applied}
    \label{fig:verticalcell1}
\end{figure}

As such, this transformation is perhaps not appropriate for producing a standardized outlier score from density metrics. For the density metric proposed here and aggregated as a geometric mean, a better choice would be something like:
$$\text{score} = -\log{(\prod_{i=1}^n d_i)^{\frac{1}{n}}} = -\frac{1}{n} \sum_{i=1}^n \log{d_i}$$
With a value of zero denoting a natural threshold between inliers and outliers. This metric would not lie between zero and one, but its distribution shape would look more similar to that of the standardized depth-based scores:
\begin{figure}[H]
\centerline{\includegraphics[totalheight=5cm]{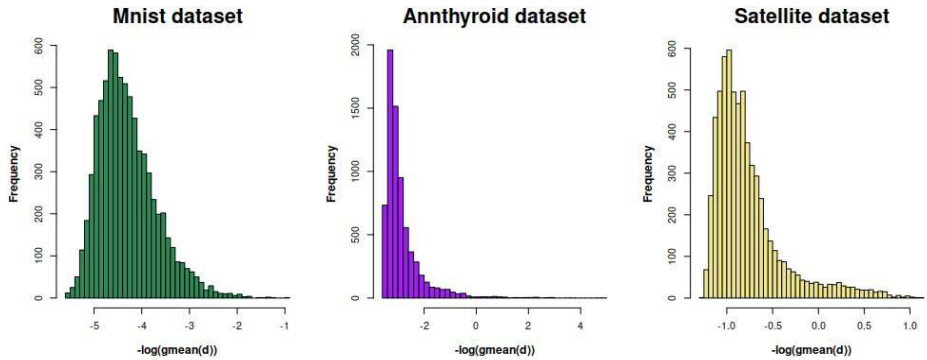}}
    \caption{Distribution of proposed outlier score under density metric}
    \label{fig:verticalcell1}
\end{figure}

For the adjusted density, using the same transformation as for isolation depth and with the same standardizing constant seems to produce a similar distribution, despite its expected value being different:

\begin{figure}[H]
\centerline{\includegraphics[totalheight=5cm]{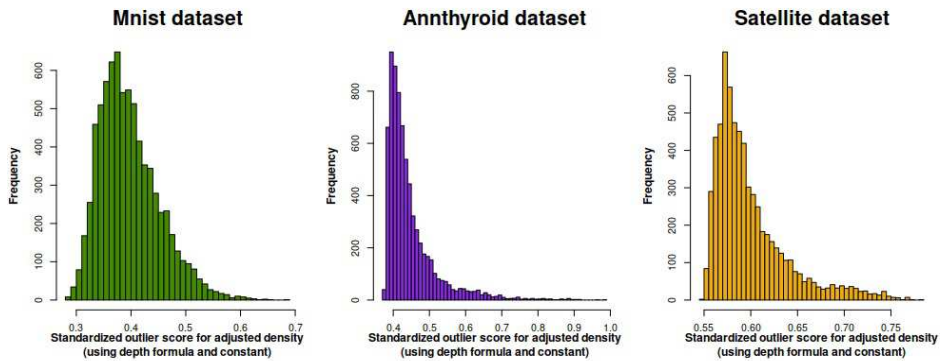}}
    \caption{Distribution of standardized outlier score for "adjusted density" using formula and constant from isolation depth}
    \label{fig:verticalcell1}
\end{figure}

For the "boxed density" metric, the score distributions look reasonable as-is in these datasets, albeit with lower scores in this distribution denoting stronger outliers as opposed to the other metrics. It unfortunately does not seem to have a common natural threshold across datasets as the other metrics do.

\section{Conclusions}

This work introduced and evaluated new density-based metrics for calculation of outlier scores from isolation forest models, which take into account the relationship between points assigned to a node branch and fraction of the feature space taken by the branch in a way which isolation depth doesn't.

These density-based metrics led to improved results in many public datasets for outlier detection across different variations of the original isolation forest model, being particularly helpful in the presence of categorical variables split through a one-vs-rest rule.

Compared to depth-based metrics, density-based metrics were found to have a larger degree of variability across trees, in some situations requiring an increased amount of trees for converge of results compared to isolation depth, and in many cases benefiniting from being aggregated through a geometric mean instead of an arithmetic mean.

New standardizing transformations were also proposed for these metrics, making them reach distributions of outlier scores that look similar to those from the the depth-based standardized score in \cite{iso}, albeit with different ranges.

\bibliographystyle{plain}
\bibliography{dens}

\end{document}